\documentclass[conference]{IEEEtran}
\IEEEoverridecommandlockouts
\usepackage{cite}
\usepackage{tcolorbox}
\usepackage[hidelinks]{hyperref}
\usepackage{amsmath,amssymb,amsfonts}
\usepackage{enumitem}
\usepackage{algorithmic}
\usepackage{array}
\usepackage{graphicx}
\usepackage{textcomp}
\usepackage{xcolor}
\usepackage{float}
\usepackage[verbose]{placeins}
\usepackage[caption=false]{subfig}
\def\BibTeX{{\rm B\kern-.05em{\sc i\kern-.025em b}\kern-.08em
    T\kern-.1667em\lower.7ex\hbox{E}\kern-.125emX}}
\begin{document}

\title{Large Language Model-Based Benchmarking Experiment Settings for Evolutionary Multi-Objective Optimization
\thanks{This work was supported by National Natural Science Foundation of China (Grant No. 62250710163, 62376115), Guangdong Provincial Key Laboratory (Grant No. 2020B121201001) (\textit{Corresponding author: Hisao Ishibuchi})}
 }

 \author{\IEEEauthorblockN{Lie Meng Pang, Hisao Ishibuchi}
 \IEEEauthorblockA{\textit{Guangdong Provincial Key Laboratory of Brain-inspired Intelligent Computation, } \\
 \textit{Department of Computer Science and Engineering,}\\
 \textit{Southern University of Science and Technology, }\\
  Shenzhen 518055, China \\
 panglm@sustech.edu.cn, hisao@sustech.edu.cn}
}

\maketitle

\begin{abstract}
When we manually design an evolutionary optimization algorithm, we implicitly or explicitly assume a set of target optimization problems. In the case of automated algorithm design, target optimization problems are usually explicitly shown. Recently, the use of large language models (LLMs) for the design of evolutionary multi-objective optimization (EMO) algorithms have been examined in some studies. In those studies, target multi-objective problems are not always explicitly shown. It is well known in the EMO community that the performance evaluation results of EMO algorithms depend on not only test problems but also many other factors such as performance indicators, reference point, termination condition, and population size. Thus, it is likely that the designed EMO algorithms by LLMs depends on those factors. In this paper, we try to examine the implicit assumption about the performance comparison of EMO algorithms in LLMs. For this purpose, we ask LLMs to design a benchmarking experiment of EMO algorithms. Our experiments show that LLMs often suggest classical benchmark settings: Performance examination of NSGA-II, MOEA/D and NSGA-III on ZDT, DTLZ and WFG by HV and IGD under the standard parameter specifications.     
\end{abstract}


\section{Introduction}
Automated algorithm design has been actively studied in the field of evolutionary multi-objective optimization (EMO) \cite{autoDesign, Auto_EMO, onlinehh_2020, onlinehh_2023, auto_MPSO, offline_EMO}. Those studies can be categorized into online approaches and offline approaches. The goal of online approaches is to find a good non-dominated solution set for a given problem by adjusting an EMO algorithm in an online manner. The goal of offline approaches is to design a high-performance EMO algorithm for a given set of multi-objective optimization problems. Similar approaches are algorithm selection where an appropriate EMO algorithm from a pre-specified algorithm pool is selected for a given multi-objective problem \cite{autoselection_emo}. A classifier such as a decision tree and a multilayer feed-forward neural network is trained based on a given set of multi-objective optimization problems. In the offline automated algorithm design and algorithm selection, it is important to use an appropriate set of multi-objective problems for EMO algorithm design and classifier design. This is because the performance of EMO algorithms (and almost all optimization algorithms) depends on the problem. It is well known that the performance evaluation results strongly depend on test problems \cite{TEVC2017}, performance indicators \cite{Ishibuchi_specifyHV} and various parameter specifications such as the termination condition and the population size \cite{CIM2022}.

Recently, the use of large language models (LLMs) in the design of new algorithms has been actively examined in the field of evolutionary computation \cite{LLaMEA, LLMSurvey_TEVC}. In the EMO field, LLMs have also been used as black-box search operators to enhance the performance of EMO algorithms \cite{LLMEMO-liufei, LLEMO-3,LLEMO-constrained}. In \cite{LLMEMO-KC}, a new LLM-based framework was proposed to autonomously design evolutionary operators for solving multi-objective optimization problems. These studies have shown promising results and suggested the potential of using LLMs to design new EMO algorithms.

In general, if we want to design an efficient EMO algorithm for a specific application problem, we will ask LLMs to design an efficient EMO algorithm for the given application problem under specific conditions (e.g., available computation time). On the contrary, if we want to design a general-purpose EMO algorithm which is applicable to a wide variety of multi-objective problems under various conditions, we will ask LLMs to design an efficient EMO algorithm without giving any particular settings for performance evaluation of EMO algorithms. Then, LLMs will design a high-performance general-purpose EMO algorithm. In this case, our question is: What implicit assumptions are used in LLMs to evaluate EMO algorithms? 

\begin{figure*}
    \centering
    \includegraphics[width=0.9\linewidth]{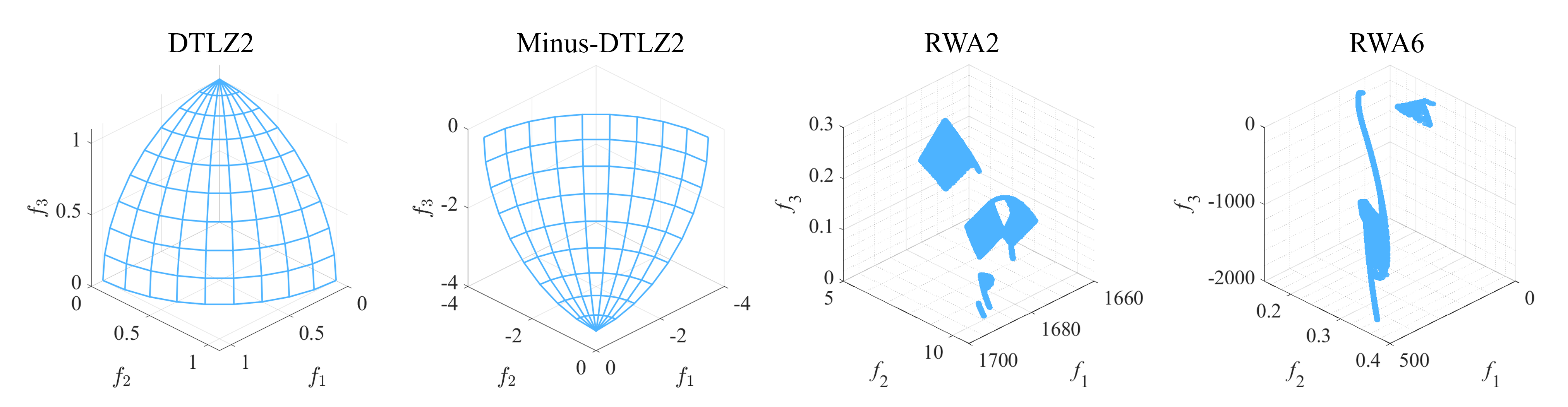}
    \caption{Pareto fronts of the DTLZ2, Minus-DTLZ2, RWA2, and RWA6 problems.}
    \label{fig:PF shapes}
\end{figure*}

In order to address this question, we use LLMs to design a benchmarking experiment for evaluating the performance of EMO algorithms. Specifically, we ask LLMs to do the following:
\begin{enumerate}
    \item Choice of EMO algorithms to be evaluated.
    \item Choice of test problems.
    \item Choice of performance indicators including the parameter specifications in each selected performance indicator.
    \item Parameter specifications in the selected EMO algorithms such as the termination condition, the population size, and the mutation and crossover probabilities. 
\end{enumerate}

By iterating this request, we obtain a number of benchmarking experiment settings. Then we analyze the obtained settings (e.g., we examine which EMO algorithms and test problems are suggested by LLMs). Our computational experiments show that somewhat classical benchmarking experimental settings are suggested. A typical suggestion is to examine the performance of NSGA-II\cite{NSGAII}, MOEA/D\cite{MOEAD},  SMS-EMOA\cite{SMSEMOA} and NSGA-III \cite{nsga3_p1} on ZDT \cite{ZDT}, DTLZ\cite{DTLZ}, WFG\cite{WFG} using the hypervolume indicator with a reference point $\textbf{r}=(r, r, \ldots, r)$ where $r=$ 1.1, and the IGD indicator with a reference point set of 10,000  uniformly generated points on the Pareto front. In this paper, we report which EMO algorithms (test problems, performance indicators) are frequently suggested by two LLMs (i.e., ChatGPT\cite{chatgpt} and DeepSeek\cite{Deepseek}) for the benchmarking of EMO algorithms. We also report the suggested specifications for EMO algorithms (e.g., population size, crossover operator, crossover probability, mutation operator, mutation probability) and performance indicators (e.g., reference point for HV and reference point set for IGD).

This paper is organized as follows. Section II provides a brief overview of the key factors influencing the performance of EMO algorithms. Section III describes the two LLMs and the methodology used in this study. Section IV presents the results and analyzes the responses provided by the LLMs. Finally, Section V concludes the paper.

\begin{figure*}
    \centering
    \includegraphics[width=1\linewidth]{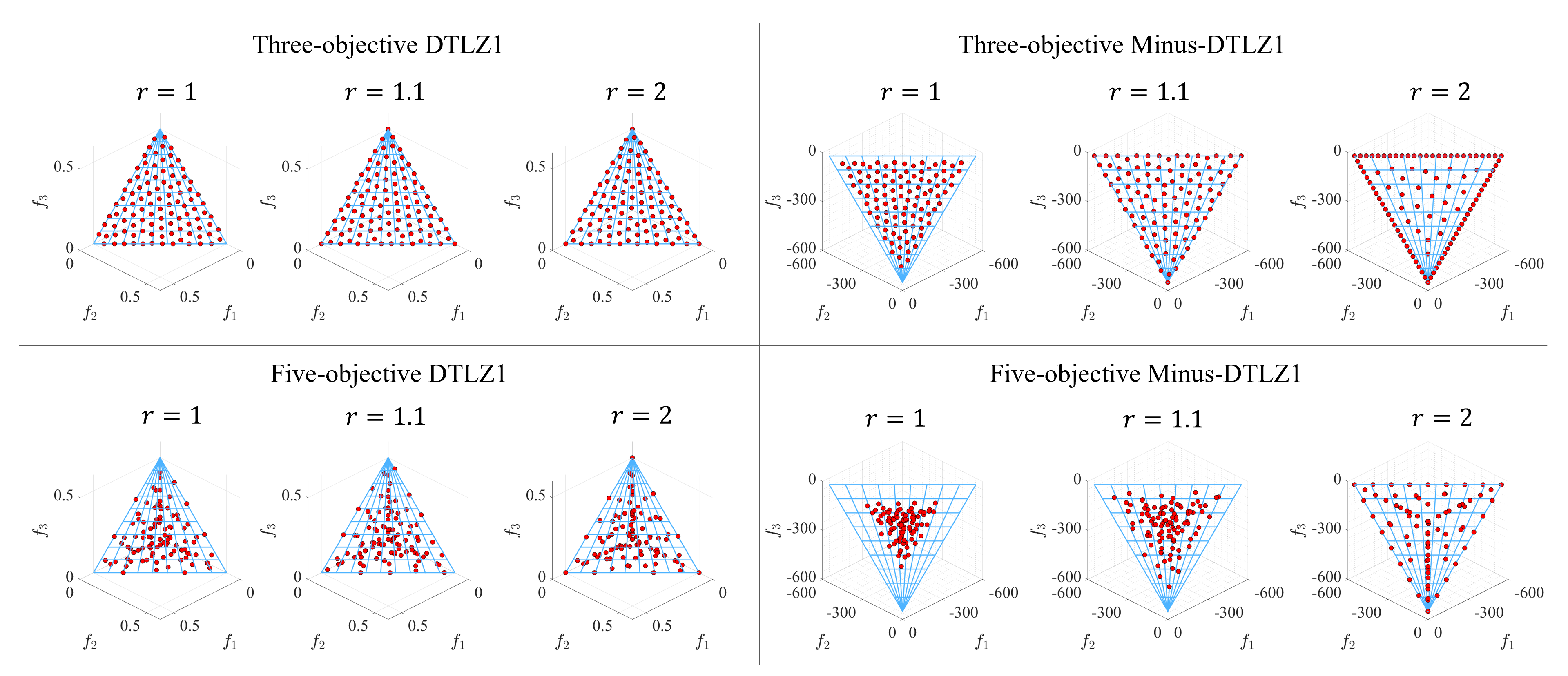}
    \caption{(Near) Optimal HV distribution for three- and five-objective DTLZ1 and Minus-DTLZ1 with different specifications of reference point. For the five-objective problems, the solutions are projected into the $f_1-f_2-f_3$ space for better visualization.}
    \label{fig:optimalHV}
\end{figure*}

\section{Performance Evaluation of EMO Algorithms}
In this section, we explain several important factors that have significant effects on the performance of EMO algorithms. These factors are considered in the design of a benchmarking experiment for EMO algorithms.

\subsection{EMO Algorithms}
When a new EMO algorithm is proposed, its performance is typically evaluated by comparing it with other EMO algorithms. The new algorithm is usually designed to outperform the selected ones. Thus, the choice of algorithms to be compared is important. It significantly affects the performance evaluation results. However, with more than a hundred EMO algorithms proposed in the literature, it is very difficult to choose their appropriate subset for performance evaluation in a convincing manner. A common approach is to select the most well-known algorithms along with several state-of-the-art algorithms. Based on the number of citations of their original papers (as shown in Table I), the following EMO algorithms are frequently used in the literature: NSGA-II, SPEA2, MOEA/D, and NSGA-III.

\begin{table}[h]
    \centering
        \caption{Number of citations of the original paper for each EMO algorithm on Google Scholar (accessed on January 16, 2025)}
    \begin{tabular}{cc}
        \hline
        \textbf{EMO algorithm} & \textbf{Number of citations} \\
        \hline
        NSGA-II & 59630\\
        \hline
        SPEA2 & 10540 \\
        \hline
        MOEA/D & 9799 \\
        \hline
        NSGA-III & 6336\\
        \hline
    \end{tabular}
    \label{tab:popular_algorithms}
\end{table}

\subsection{Test Problems}
Test problems also play a critical role in the performance comparison of EMO algorithms. The choice of test problems can lead to completely different evaluation results. Historically, ZDT \cite{ZDT} and DTLZ \cite{DTLZ} were frequently used in the 2000s. In the 2010s, WFG \cite{WFG} became commonly used alongside ZDT and DTLZ. In the 2020s, various test problems, such as Minus-DTLZ and Minus-WFG \cite{TEVC2017}, MaF \cite{MaF}, and real-world problems \cite{RE, DDMOP, RWA}, were increasingly used in the EMO community in addition to ZDT, DTLZ, and WFG. This is because some recent studies have pointed out that the DTLZ and WFG test problems are too regular and unrealistic \cite{PF_regular}. As shown in Fig. 1, the Pareto front of DTLZ2 has a triangular simplex shape. In contrast, the Pareto front of Minus-DTLZ2 is inverted triangular, which is more realistic as explained in \cite{PF_regular}. The Pareto fronts of the two real-world problems (i.e., RWA2 and RWA6) have irregular shapes.

\subsection{Performance Indicators}
As overall performance indicators, the hypervolume (HV) \cite{Zitzler_98} and inverted generational distance (IGD) \cite{igd} indicators have been frequently used. A reference point is needed for HV calculation, and a reference point set is needed for IGD calculation. A slightly worse point than the nadir point in the objective space is usually used as the reference point for HV. The reference point set for IGD is usually a  set of a large number of uniformly sampled points on the Pareto front (e.g., 10,000 points). In general, the HV-based and IGD-based performance comparison results depend on the specification of the reference point and the reference point set, respectively.

One interesting observation in the literature \cite{Ishibuchi_specifyHV} is that the effect of the reference point on the HV optimal solution distribution is small for triangular Pareto fronts but large for inverted triangular Pareto fronts. Fig. 2 shows the (near) HV-optimal distributions of solutions obtained by SMS-EMOA on the three-objective and five-objective DTLZ1 and Minus-DTLZ1 for three different reference point specifications. The reference point is specified as $\textbf{r}=(r, r, \ldots, r)$ where $r=$ 1.1, 1.5, and 2. We can observe the following in Fig. 2:
\begin{enumerate}[label=(\roman*)]
\item The reference point specification is not very important for triangular Pareto fronts. For example, the HV-optimal solution distribution of the three-objective DTLZ1 problem can uniformly cover the entire Pareto front when any reference point with a value worse than the nadir point (such as $r=$ 1.1 and 2) is used. 
\item The reference point specification is very important for irregular Pareto fronts (e.g., inverted triangular shape). For example, $r=$ 1.1 is suitable for the three-objective Minus-DTLZ1 problem as the HV-optimal solution distribution covers the entire Pareto front. When $r=$ 2, many solutions lie on the boundary of the Pareto front, and only a small number of solutions are inside. That is, the HV-optimal solution distribution is sensitive to the reference point specification when the Pareto front shape is irregular.
\item An appropriate specification of the reference point depends on the number of objectives. $r=$ 1.1 is suitable for three-objective problems but it is too small (i.e., too close to the Pareto front) in the case of five-objective problems (the optimal distribution of solutions does not spread over the entire Pareto front when $r=$ 1.1). It seems that $r=$ 2 is a better specification than $r=$ 1.1 for the five-objective problems since boundary solutions are obtained.   
\end{enumerate}

\section{Methodology}
Generally, large language models (LLMs) are deep neural networks that have been specifically trained on massive amounts of text data to perform tasks related to human natural language usage \cite{LLM}. LLMs have been widely used in a variety of applications such as text generation, code generation, code completion, biomedicine \cite{LLM_survey}. In the field of EMO, the use of LLMs as evolutionary optimizers and automatic designers of evolutionary operators has been explored in some studies \cite{LLMEMO-liufei, LLEMO-3,LLEMO-constrained,LLMEMO-KC}. These studies implicitly assume that LLMs possess rich domain knowledge that can be used to assist or accelerate evolutionary processes. Thus, a natural question arises from these studies: What kind of implicit knowledge do LLMs use to assist the search process or to help design and generate new EMO algorithms?

In this study, we attempt to answer this question by analyzing the responses provided by two LLMs, namely ChatGPT-4o \cite{chatgpt} and DeepSeek-V3 \cite{Deepseek}. We choose these two LLMs based on the following considerations:
\begin{itemize}
    \item The ChatGPT series is one of the most well-known closed-source LLMs developed by OpenAI, and they have been used in \cite{LLMEMO-liufei, LLEMO-3,LLEMO-constrained,LLMEMO-KC}. ChatGPT-4o is the company's latest flagship model. Thus, we consider it as a representative closed-source LLM in our study.
    \item  DeepSeek-V3 is a fully open-source LLM developed by DeepSeek, which has recently demonstrated remarkable performance in terms of language understanding and reasoning capabilities. It was reported that its performance is comparable to leading closed-source LLMs, such as the ChatGPT series \cite{Deepseek_report}. Thus, we use DeepSeek-V3 as a representative open-source LLM in our study.  
\end{itemize}

In our experiment, we assigned the LLMs the role of an expert in the EMO field and posed questions related to the benchmarking of EMO algorithms in a zero-shot manner. For each question, we collected 10 responses from the LLMs to account for the stochastic nature in their outputs. For parameters such as temperature in the LLMs, we used the default settings provided by each LLM.

\section{Results}
In this section, we present the results obtained from the two LLMs for a series of questions related to the benchmarking of EMO algorithms. First, we provide the following prompt to the LLMs to obtain general suggestions for benchmarking EMO algorithms: 

\begin{tcolorbox}[colback=gray!8, colframe=gray!40, arc=3mm, boxrule=0.3mm,width=8.8cm,fontupper=\small\itshape]
\textit{You are an expert in evolutionary multi-objective optimization (EMO). What are your suggestions for benchmarking EMO algorithms? Keep your response short and concise.}
\end{tcolorbox}

Among the 10 responses provided by the two LLMs, the following suggestions are almost always included: \begin{itemize} 
\item Use diverse test problems 
\item Include real-world problems 
\item Use established performance metrics like HV and IGD 
\item Conduct scalability testing for both decision and objective spaces 
\item Ensure reproducibility 
\item Perform statistical validation 
\item Use visual tools like the Pareto front for interpretability 
\item Perform runtime analysis 
\item Perform parameter sensitivity analysis to ensure robustness
\end{itemize}

These suggestions are reasonable and align with the current understanding in the community regarding the benchmarking of EMO algorithms. In the subsequent subsections, we ask the two LLMs more detailed questions related to each category, including suggestions for EMO algorithms, test problems, performance indicators, and other parameter settings. 

\subsection{Suggested EMO Algorithms}
First, we asked the LLMs to suggest how many EMO algorithms should be included in a benchmarking experiment. Both LLMs suggested including five to ten EMO algorithms in the experiment. Based on this suggestion, we then asked the LLMs to suggest five EMO algorithms for benchmarking, using the following prompt: 

\begin{tcolorbox}[colback=gray!8, colframe=gray!40, arc=3mm, boxrule=0.3mm,width=8.8cm,fontupper=\small\itshape]
\textit{You are an expert in evolutionary multi-objective optimization (EMO). Please suggest five EMO algorithms for benchmarking. Keep your response short and concise.}
\end{tcolorbox}

Table II shows the percentage of times each algorithm was included in the responses given by the LLMs. Both ChatGPT-4o and DeepSeek-V3 consistently suggested NSGA-II, MOEA/D, SPEA2, and NSGA-III for benchmarking in all of their 10 responses. RVEA was suggested by ChatGPT-4o in 60\% of its responses, while DeepSeek-V3 did not suggest it in any of its 10 responses. In contrast, DeepSeek-V3 suggested HypE and IBEA in 60\% and 40\% of its responses, respectively. 

\begin{table}
    \centering
    \caption{Percentage of times each algorithm was suggested by ChatGPT-4o and DeepSeek-V3 in response to the request for five EMO algorithms (based on 10 responses per model).}
    \begin{tabular}{ccc}
    \hline
          Algorithm& ChatGPT-4o (\%) & DeepSeek-V3 (\%)\\
    \hline
         NSGA-II & 100 & 100\\
    \hline
         MOEA/D & 100 & 100\\
    \hline
         SPEA2 & 100 & 100\\
     \hline
        NSGA-III & 100 & 100\\
    \hline 
        RVEA & 60 & 0\\
    \hline 
        R-NSGA-II & 20 & 0\\
    \hline
        HypE & 20 &60\\
    \hline
        IBEA & 0 & 40\\
    \hline
    \end{tabular}
    
    \label{tab:five_algorithms}
\end{table}

\begin{table}
    \centering
    \caption{Percentage of times each algorithm was suggested by ChatGPT-4o and DeepSeek-V3 in response to the request for five state-of-the-art EMO algorithms (based on 10 responses per model).}\begin{tabular}{ccc}
    \hline
         Algorithm & ChatGPT-4o (\%) & DeepSeek-V3 (\%)\\
    \hline
         RVEA& 100 & 40\\
    \hline 
        NSGA-III & 80 & 100\\
    \hline
        MOEA/D-DRA & 80 & 0\\
    \hline
        HypE & 80 & 100\\
    \hline
         AGE-II& 60 & 0\\
    \hline
        U-NSGA-III& 40 & 0\\
    \hline
        SMS-EMOA & 30 & 30 \\
    \hline
        MOEA/D-DE& 20 & 70\\
    \hline 
        CMA-ES/MA& 10 & 0\\
    \hline
        MOEA/DD& 0 & 30\\
    \hline
        LMEA&0 & 40\\
    \hline
        AR-MOEA &0 &90\\
    \hline 
    \end{tabular}
    
    \label{tab:five_state_of_art}
\end{table}

Next, we slightly modified the prompt by asking the LLMs to suggest ``five state-of-the-art EMO algorithms" for benchmarking. Table III shows the percentage of times each algorithm was included in the responses given by the LLMs. For ChatGPT-4o, RVEA was always suggested as a state-of-the-art EMO algorithm, whereas NSGA-III and HypE were always suggested by DeepSeek-V3 in all 10 responses. Based on Table III, it seems that ChatGPT-4o and DeepSeek-V3 have somewhat different knowledge about state-of-the-art EMO algorithms. For example, MOEA/D-DRA was suggested in 80\% of the responses by ChatGPT-4o, while it was not suggested by DeepSeek-V3. AR-MOEA was suggested by DeepSeek-V3 in 90\% of the responses, while it was not suggested by ChatGPT-4o. 

\subsection{Suggested Test Problems}
To understand the implicit knowledge in the LLMs about the test problems for EMO, we asked the following question:

\begin{tcolorbox}[colback=gray!8, colframe=gray!40, arc=3mm, boxrule=0.3mm,width=8.8cm,fontupper=\small\itshape]
\textit{You are an expert in evolutionary multi-objective optimization (EMO). Please suggest test problems for benchmarking EMO algorithms. Keep your response short and concise.}
\end{tcolorbox}

\begin{table}
    \centering
    \caption{Percentage of times each problem was suggested by ChatGPT-4o and DeepSeek-V3 in response to the request for suggesting test problems for benchmarking (based on 10 responses per model).}
    
    \begin{tabular}{ccc}
    \hline
         Problems & ChatGPT-4o (\%) & DeepSeek-V3 (\%)\\
    \hline
         ZDT suite& 100 & 100\\
    \hline 
        DTLZ suite & 100 & 100\\
    \hline
        WFG suite & 100 & 100\\
    \hline
        UF suite & 70 & 100\\
    \hline
        Real-world problems& 70 & 0\\
    \hline
        CEC competition problems& 50 & 0\\
    \hline
        CF suite & 10 & 0 \\
    \hline
        LZ suite & 0 & 100\\
    \hline 
    \end{tabular}
    
    \label{tab:test problems}
\end{table}

Table \ref{tab:test problems} shows the percentage of times each test problem was included in the responses given by the LLMs. The LLMs suggested a number of test suites in each of the 10 responses, as listed in Table \ref{tab:test problems}. All the ZDT, DTLZ, and WFG test suites are included in all the responses from both LLMs. This is not surprising since these three test suites are the most well-known in the EMO field and have been used in many studies. It is also interesting to note that there is no variation in the responses given by DeepSeek-V3: All 10 responses consistently include the ZDT, DTLZ, WFG, UF, and LZ suites as the suggested test problems. For ChatGPT-4o, it suggested real-world problems in 70\% of its responses, whereas DeepSeek-V3 did not suggest any real-world problems. The real-world problems suggested by ChatGPT-4o include: portfolio optimization, vehicle routing, engineering design (e.g., gear design, truss optimization), knapsack problems, scheduling problems, and network routing problems. 

Next, we slightly modified the prompt by specifically asking the LLMs to suggest ``test suites" for benchmarking. The results are shown in Table \ref{tab:test suites}. In general, the results are very similar to the results in Table \ref{tab:test problems}. For DeepSeek-V3, its responses are the same as those in Table \ref{tab:test problems} where the ZDT, DTLZ, WFG, UF, and LZ suites were always suggested. The responses from ChatGPT-4o were slightly different from those in Table \ref{tab:test problems}, where the frequency of the CEC competition problems and CF suites increased from 50\% and 10\% to 70\% and 50\%, respectively. Additionally, MaF was included in the response, while real-world problems were suggested less often in comparison to Table \ref{tab:test problems}.

Finally, we ask the LLMs to suggest ``only one test suite" for benchmarking EMO algorithms. Both the LLMs consistently suggested the DTLZ suite in all their responses. 

\begin{table}
    \centering
    \caption{Percentage of times each problem suite was suggested by ChatGPT-4o and DeepSeek-V3 in response to the request for suggesting test suites for benchmarking (based on 10 responses per model).}\begin{tabular}{ccc}
    \hline
         Test suite & ChatGPT-4o (\%) & DeepSeek-V3 (\%)\\
    \hline
         ZDT suite& 100 & 100\\
    \hline 
        DTLZ suite & 100 & 100\\
    \hline
        WFG suite & 100 & 100\\
    \hline
        UF suite & 70 & 100\\
    \hline
        CEC competition problems& 70 & 0\\
    \hline
        CF suite & 50 & 0 \\
    \hline 
        MaF suite & 30 & 0\\
    \hline
        Real-world problems& 20 & 0\\
    \hline
        LZ suite & 0 & 100\\
    \hline 
    \end{tabular}
    
    \label{tab:test suites}
\end{table}

\begin{figure*}
    \centering
    \includegraphics[width=0.6\linewidth]{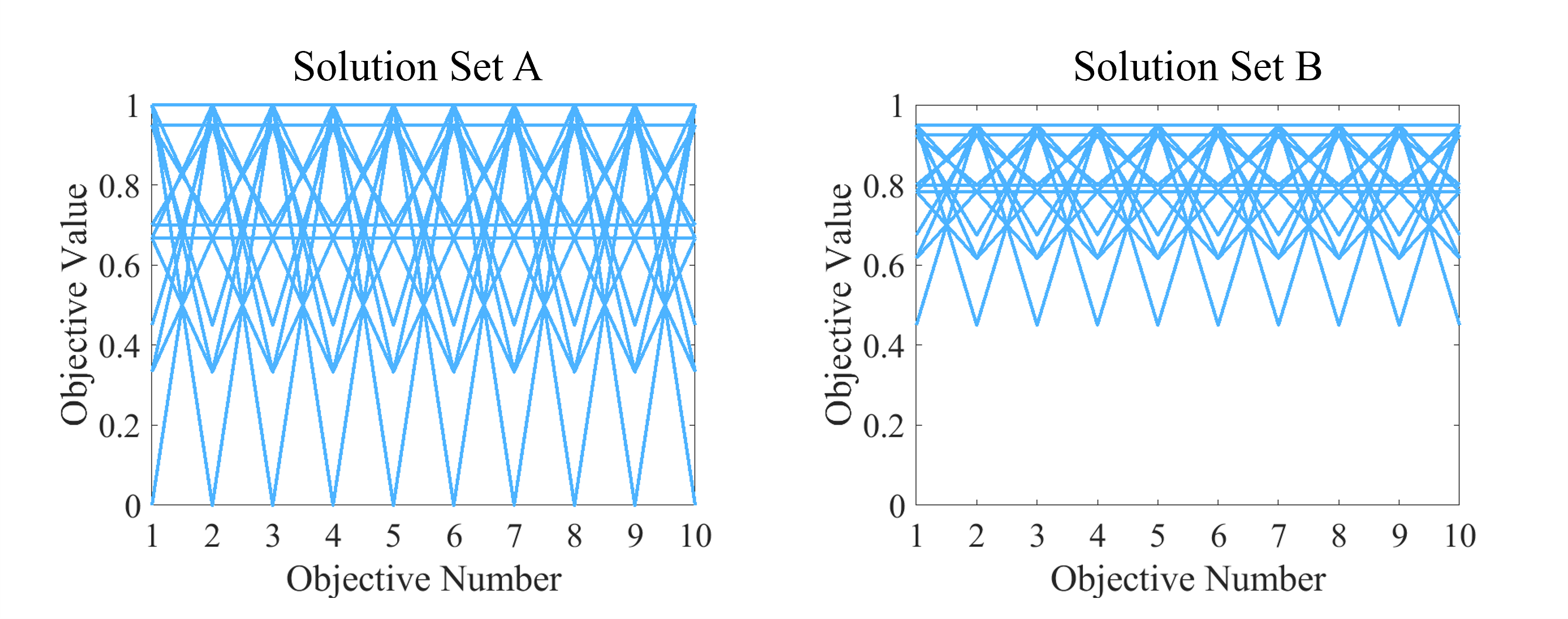}
    \caption{Two Pareto-optimal solution sets with different solution distributions for the normalized ten-objective Minus-DTLZ1 problem, each containing 275 solutions.}
    \label{fig:two solution sets}
\end{figure*}

\subsection{Suggested Performance Indicators}
In this subsection, we focus on the performance indicators for evaluating EMO algorithms. We first asked the LLMs to suggest performance indicators for benchmarking EMO algorithms, using the following prompt:

\begin{tcolorbox}[colback=gray!8, colframe=gray!40, arc=3mm, boxrule=0.3mm,width=8.8cm,fontupper=\small\itshape]
\textit{You are an expert in evolutionary multi-objective optimization (EMO). Please suggest performance indicators for benchmarking EMO algorithms. Keep your response short and concise.}
\end{tcolorbox}

Table \ref{tab:performance indicator} shows the percentage of times each performance indicator was included in the responses given by the LLMs. For ChatGPT-4o, HV and IGD were always suggested, while $\epsilon$-indicator and spread were included in 90 \% and 80\% of the responses, respectively. For DeepSeek-V3, HV, IGD, $\epsilon$-indicator, and spread were always included in the responses. Again, we observe that the responses from ChatGPT-4o and DeepSeek-V3 are slightly different. Compared to DeepSeek-V3, ChatGPT-4o suggested more indicators, including computational time, convergence metric, and R2. 

\begin{table}
    \centering
    \caption{Percentage of times each performance indicator was suggested by ChatGPT-4o and DeepSeek-V3 in response to the request for suggesting performance indicators for benchmarking (based on 10 responses per model).}\begin{tabular}{ccc}
    \hline
        Performance indicator & ChatGPT-4o (\%) & DeepSeek-V3 (\%)\\
    \hline
        HV & 100 & 100\\
    \hline 
        IGD & 100 & 100\\
    \hline
        $\epsilon$-indicator & 90 & 100\\
    \hline
        Spacing & 80 & 40\\
    \hline
        Computational time& 50 & 0\\
    \hline
       Spread & 30 & 100 \\
    \hline 
        GD & 30 & 60\\
    \hline
       Convergence metric& 10 & 0\\
    \hline
        R2 & 10 & 0\\
    \hline 
    \end{tabular}
    
    \label{tab:performance indicator}
\end{table}

Next, we asked the LLMs to suggest ``only a single performance indicator" for benchmarking EMO algorithms using the following prompt: 
\begin{tcolorbox}[colback=gray!8, colframe=gray!40, arc=3mm, boxrule=0.3mm,width=8.8cm,fontupper=\small\itshape]
\textit{You are an expert in evolutionary multi-objective optimization (EMO). I want to use a single performance indicator for benchmarking EMO algorithms. Could you suggest an appropriate one? Keep your response short and concise.}
\end{tcolorbox}
Both LLMs consistently suggested HV as the most appropriate performance indicator in all 10 responses. The LLMs included the following reason for the suggestion of HV in their responses: ``HV can comprehensively assess both the convergence and diversity of a solution set simultaneously". This aligns with knowledge in the EMO community, where HV is primarily used because it is the only Pareto-compliant indicator. However, the explanation about ``Pareto-compliant" was not included in the responses.

\subsubsection{Suggested reference point for HV}
Since the calculation of HV requires the specification of a reference point, we further asked the LLMs to provide suggestions on how to specify a reference point for benchmarking EMO algorithms. In general, both LLMs suggested choosing a point slightly worse than the nadir point of the problem, with the common practice of adding a small offset (e.g., 10\% or 20\%) to the nadir point. Normalization of the objective space for HV calculation was suggested in some of their responses. In that case, a reference point of (1.1, \ldots, 1.1) was suggested. 

Next, we asked the LLMs to suggest the reference point specification for two-objective, three-objective, five-objective and ten-objective problems after normalizing the objective space so that the ideal point is (0, \ldots, 0) and the nadir point is (1, \ldots, 1). The population size for each number of objectives is also provided in the prompt. For two-objective, three-objective, five-objective, and ten-objective problems, with the population sizes are about 100, 100, 120, and 275, respectively. The prompt for asking the reference point specification for two-objective problems is shown as follows: 

\begin{tcolorbox}[colback=gray!8, colframe=gray!40, arc=3mm, boxrule=0.3mm,width=8.8cm,fontupper=\small\itshape]
\textit{You are an expert in evolutionary multi-objective optimization (EMO). For hypervolume calculation, how to specify a reference point for a 2-objective problem after the normalization of the objective space so that the ideal point is (0, 0, \ldots, 0) and the nadir point is (1, 1, \ldots, 1). The population size is about 100. Keep your response short and concise.}
\end{tcolorbox}

In all responses given by the two LLMs, the suggested reference points were (1.1, \ldots, 1.1) for two-objective, three-objective, five-objective, and ten-objective problems. Whereas (1.1, \ldots, 1.1) is a frequently-used reference point, it is not always appropriate especially for many-objective problems, as we have explained in Section II.C. It should also be noted that it is needed to use a different reference point specification depending on the population size and the number of objectives, as highlighted in a recent study \cite{Ishibuchi_specifyHV}. In order to obtain more reasonable comparison results using HV, the following reference point specification method in the normalized objective space was proposed in \cite{Ishibuchi_specifyHV}: 
\begin{equation}
    r = 1 + 1/H.
\end{equation}
The value of $H$ is specified based on the following formulation: 
\begin{equation}
    C_{m-1}^{H+m-1} \leq N < C_{m-1}^{H+m},
\end{equation}
where $m$ is the number of objectives and $N$ is the population size. As an example, given a solution set with 275 solutions (i.e., $N=275$) for a ten-objective problem, we can obtain $H=$ 3, and the reference point should be specified as (4/3, \ldots, 4/3). 

Fig. \ref{fig:two solution sets} provides an example that demonstrates the importance of using an appropriate reference point specification based on the population size. In Fig. \ref{fig:two solution sets}, both Solution Set A and Solution Set B are Pareto-optimal solution sets, each containing 275 solutions. By visual evaluation, it is clear that A is better than B, since A covers the entire Pareto front, while B covers only the inside solutions. However, using the reference point (1.1, \ldots, 1.1), B is evaluated as better than A: \text{HV(B)} $= 2.3975 \times 10^{-6} >$ HV(A) $= 1.3264 \times 10^{-6}$. When the reference point is specified as (4/3, \ldots, 4/3), A is evaluated as better than B: HV(A) $=$ 0.0054 $>$ HV(B) $=$ 0.0035. Thus, the reference point specification proposed in \cite{Ishibuchi_specifyHV} leads to more intuitively acceptable comparison results, whereas such a specification was not suggested by the two LLMs.

\subsubsection{Suggested reference point set for IGD} 
In addition to HV, IGD is another frequently-used indicator in the EMO field. A reference point set is required for the calculation of the IGD value. In order to understand what are the knowledge of the LLMs regarding the specification of the reference point set for IGD, we ask the following question: 

\begin{tcolorbox}[colback=gray!8, colframe=gray!40, arc=3mm, boxrule=0.3mm,width=8.8cm,fontupper=\small\itshape]
\textit{You are an expert in evolutionary multi-objective optimization (EMO). Please suggest how to specify a reference point set for IGD calculation when benchmarking EMO algorithms. Keep your response short and concise.}
\end{tcolorbox}

In general, both LLMs include the following suggestions for specifying the IGD reference point set in their responses:
\begin{itemize}
    \item If the the true Pareto is available, use a well-distributed (uniformly-distributed) set sampled from the true Pareto front. 
    \item If the true Pareto front is not available, approximate the true Pareto front by combining non-dominated solutions from all algorithms and sampling uniformly. 
    \item Ensure the reference set is sufficiently large (e.g., 1,000 to 10,000 points) to accurately represent the Pareto front. 
\end{itemize}

Next, we asked the LLMs to suggest how many reference points are needed in the reference point set. When the prompt did not explicitly mention the number of objectives, both LLMs suggested using 1,000 to 10,000 uniformly distributed reference points on the Pareto front in most of their responses. Then, we explicitly asked the LLMs to suggest the number of reference points for two-objective, three-objective, five-objective, and ten-objective problems.

For ChatGPT-4o, the following number of reference points were suggested (summarized based on ten responses): 
\begin{itemize}
    \item Two-objective: 500 to 10,000 points
    \item Three-objective: 5,000 to 10,000 points
    \item Five-objective: 20,000 to 100,000 points
    \item Ten-objective: 500,000 to 1,000,000 or more points
\end{itemize}

For DeepSeek-V3, the following number of reference points were suggested (summarized based on ten responses): 
\begin{itemize}
    \item Two-objective: 1,000 to 10,000 points
    \item Three-objective: 10,000 to 50,000 points
    \item Five-objective: 50,000 to 100,000 points
    \item Ten-objective: 100,000 to 1,000,000 points
\end{itemize}

These responses show that both LLMs understand the need for a large number of uniformly distributed reference points for IGD calculation. They also understand the importance of scaling the number of reference points based on the number of objectives. As a result, the number of reference points is adjusted accordingly for a different number of objectives.

However, it is worth mentioning that the use of a large number of uniformly distributed reference points can lead to counterintuitive performance comparison results in many-objective cases. A solution set with smaller diversity can have a better (i.e., smaller) IGD value than a well-distributed solution set over the entire Pareto front due to the bias of the IGD indicator (i.e., no boundary solutions are included in the optimal distribution of solutions for the IGD indicators), as explained in \cite{igd_hisao}.

\subsection{Suggested Parameter Values in EMO Algorithms}
Finally, in this subsection, we aim to understand the knowledge of parameter values in EMO algorithms from the two LLMs, particularly about the population size specification, and the choice of crossover and mutation operators.

\subsubsection{Population size}
We asked the LLMs how to determine the population size for EMO algorithms and requested that they suggest the population size for two-objective, three-objective, five-objective, and ten-objective problems for benchmarking EMO algorithms, respectively. Both LLMs explained that the population size for EMO algorithms typically depends on the number of objectives and the problem's complexity. They also explained that a larger population size is needed for higher-dimensional objective spaces to help maintain diversity.

For ChatGPT-4o, the suggested population sizes are summarized based on its ten responses as follows:
\begin{itemize}
    \item Two-objective: 100 to 300 individuals 
    \item Three-objective: 120 to 500 individuals 
    \item Five-objective: 200 to 1,000 individuals 
    \item Ten-objective: 500 to 5,000 individuals 
\end{itemize}

For DeepSeek-V3, the suggested population sizes are summarized based on its ten responses as follows:
\begin{itemize}
    \item Two-objective: 100 to 200 individuals 
    \item Three-objective: 120 to 300 individuals 
    \item Five-objective: 200 to 500 individuals 
    \item Ten-objective: 300 to 1,000 individuals 
\end{itemize}

As we can see, the suggested population sizes are scaled up based on the number of objectives, with up to 5,000 suggested by ChatGPT-4o for ten-objective problems. While these suggestions seem reasonable, in the EMO community, a population size of less than 300 (with many studies using 275) is typically used for ten-objective problems. This is because using a large population size (e.g., 5,000) means a small number of generations (which often deteriorates the performance of many EMO algorithms) when the total number of solution evaluations is used as the termination condition.

\subsubsection{Crossover operators}
Next, we asked the LLMs for suggestions on selecting the crossover operator and probability. Both LLMs included the simulated binary crossover (SBX) and uniform crossover in their responses. However, ChatGPT-4o occasionally incorrectly explained that SBX is for binary-coded problems, whereas SBX is actually used for real-coded problems. In addition, ChatGPT-4o suggested other crossover operators, such as the blend crossover and differential evolution crossover, with the crossover probability between 0.8 and 1.0 in most of its responses. As for DeepSeek-V3, it consistently suggested SBX crossover (for real-coded problems) and uniform crossover (for binary-coded problems) in all ten responses, with a crossover probability of 0.9.

\subsubsection{Mutation operators}
For the choice of mutation operators and mutation probability, both LLMs consistently suggested polynomial mutation (for real-coded problems) and bit-flip mutation (for binary-coded problems) in all their responses. Additionally, ChatGPT-4o suggested other mutation operators, including Gaussian mutation and non-uniform mutation, in some of its responses. As for the mutation probability, ChatGPT-4o suggested setting it between 0.01 and 0.1 in most of its responses. For DeepSeek-V3, the mutation probability was suggested to be set as $1/n$, where $n$ is the number of decision variables.

\section{Concluding Remarks}
In order to understand the implicit knowledge used in LLMs for designing EMO algorithms, this study asked two large language models (LLMs) for their suggestions on benchmarking EMO algorithms. As shown by the results presented in this paper, the LLMs suggest historically widely-used settings for benchmarking. Whereas those settings have been used in many papers in the literature, they are not necessarily good settings in many points such as the reality of test problems and the reference point (set) specification for HV (IGD) calculation. Our results may imply that the LLM-based general purpose automated algorithm design will be adjusted to the suggested settings, which may mean that the designed algorithm works very well on the suggested test problems based on the suggested performance evaluation mechanisms. We hope that LLMs will suggest better settings in the near future.

\end{document}